\documentclass{amsart}[12 pt]
\usepackage[backend=biber,style=authoryear,maxcitenames=1]{biblatex}
\usepackage[colorlinks=true]{hyperref}
\usepackage[foot]{amsaddr}
\usepackage{latexsym}
\usepackage{amsmath,amssymb,dsfont,amsfonts}
\usepackage{mathrsfs}
\usepackage{verbatim}
\usepackage{graphicx}
\graphicspath{ {./SQM10Fig/} }
\usepackage{graphics}
\usepackage{geometry}
\usepackage{booktabs}
\usepackage{enumitem}
\usepackage{multirow}
\usepackage[skip=3ex]{subcaption}

\newtheorem{theorem}{Theorem}[section]

\newcommand{\E}{\mathbb{E}}
\newcommand{\diag}{\operatorname{diag}}

\newcommand{\zt}{z^\top}
\DeclareMathOperator*{\argmin}{arg\,min}
\DeclareMathOperator*{\prox}{prox}

\numberwithin{equation}{section}

\title{Randomized SINDy}

\author{Dorival Leão $^{1,a}$}
\address{$^1$Estatcamp Consultoria, São Carlos - SP, Brazil}
\email{$^a$leao@estatcamp.com.br}
\author{Reiko Aoki $^{2,b}$}
\address{$^2$Instituto de Ciências Matemáticas e de Computação.
Universidade de São Paulo, São Carlos - SP, Brazil}
\email{$^b$reiko@icmc.usp.br}
\author{Alberto Ohashi $^{3,c}$}
\address{$^3$Instituto de Ciências Exatas. Departamento de Matemática.
Universidade de Brasília, Brasília - DF, Brazil}
\email{$^c$ohashi@mat.unb.br}
\author{Teh Led Red $^{1,d}$}
\email{$^d$teh@estatcamp.com.br}

\usepackage[foot]{amsaddr}
\date{April 2026}

\allowdisplaybreaks

\begin{document}

\begin{abstract}
This paper presents ``randomized SINDy", a sequential machine learning algorithm designed for dynamic data that has a time-dependent structure. It employs a probabilistic approach, with its PAC learning property rigorously proven through the mathematical theory of functional analysis. The algorithm dynamically predicts using a learned probability distribution of predictors, updating weights via gradient descent and a proximal algorithm to maintain a valid probability density. Inspired by SINDy (\cite{SINDy}), it incorporates feature augmentation and Tikhonov regularization. For multivariate normal weights, the proximal step is omitted to focus on parameter estimation. The algorithm's effectiveness is demonstrated through experimental results in regression and binary classification using real-world data.
\end{abstract}
\maketitle

\section{Introduction}
Sequential machine learning represents a family of efficient and scalable algorithms
to incrementally build predictive models from a sequence of data (\cite{Rosenblatt}). Traditional machine learning algorithms have a high computational cost for retraining in the presence of new data while sequential algorithms are suitable for applications with a high volume of data sampled dynamically and quickly. Huge databases that grow dynamically over time require scalable machine learning algorithms \parencite{Kivinen, Lu, Tu}.

Traditional machine learning algorithms assume that the data is independent and identically distributed (iid). However, in many cases involving the mining of text, voice and time series data, the data in step $t$ depends on the data from the previous steps. This brings us to sequential learning algorithms. Recurrent neural networks (RNN) are a class of machine learning algorithms used to treat data correlated in time as time series. However, RNN have a high computational cost, which makes them difficult to apply in the sequential learning scenario \parencite{Sutskever, Elsayed}.

One class of sequential learning algorithms that is widely used is the ``Linear Sequential Learning" class of algorithms \parencites{Rosenblatt, Crammer_06, Dredze},  which establishes a linear predictive model. The limitation of this class of algorithms is that the model is linear, which is not valid for most applications. This issue motivated the development of ``kernel-based sequential learning" models \parencite{Kivinen},  which use kernel-based predictive models. In general, kernel-based sequential learning models are not scalable as the set of support vectors used in the predictive model is unlimited, which can be a problem when dealing with large volumes of data. A simple strategy is to limit the size of the set of support vectors \parencite{Crammer_03}. With this strategy, we can dispense with observations that are currently irrelevant, but could play a fundamental role in the predictive model a few steps down the line. An alternative is to use techniques related to ``Random Fourier features" \parencite{Rahimi} or fast food techniques \parencite{Lu} to approximate the original kernel function \parencite{Lu, Tu}. These methods map the data to the random-feature space in which they apply stochastic gradient algorithms. However, in order to obtain good approximations, a high number of random features is required, which can cause computational problems.

Furthermore, these general predictive models do not provide us with a measure of confidence in the forecasts they make. In various applications, such as demand forecasting, it is important to establish a measure of confidence in the forecasts so that we can develop strategies for stock control. One strategy for developing a measure of confidence in forecasts is to use Bayesian techniques to quantify the uncertainty related to the model and estimates \parencite{Rasmussen, Burnaev}. Another alternative is to use ``Conformal Prediction" developed by \textcite{Vovk}, which uses the assumption of independent observations to construct a set in the space of labels with a given probability \parencite{Burnaev}. Although the conformal prediction technique is distribution-free, it assumes that the data is independent.

The main objective of this work is to develop a class of algorithms that works incrementally that is suitable for large volumes of data updated online and has a time-dependent structure.  In addition, we will propose a measure of confidence for the predictions made by the algorithm. To do this, we develop an extension of the class of SINDy models proposed by \cite{SINDy}, incorporating random coefficients that are estimated and updated online. In this case, we leverage online convex optimization techniques to sequentially update the coefficients and to obtain probabilistic coverage guarantees for the prediction estimate.

This paper is organized as follows: in \autoref{sec:RSindy}, we introduce the proposed algorithm, randomized SINDy, and demonstrates its PAC learning properties through rigorous analysis. In \autoref{sec:experiments}, we present experimental results, starting with simple simulations to verify the effectiveness of the proposed approach, followed by applications on two real-world datasets. Finally, \autoref{sec:conclusion} concludes with a discussion of the results and future research directions.

\section{Randomized SINDy}\label{sec:RSindy}

We consider the problem of sequential learning by the following online convex optimization setting.  At the initial step, we observe $x_1 \in \mathbb{R}^k$. Then the learner takes an initial action $A_1$ which is a probability distribution over the space of labels $\mathbb{R}^m$ as follows

\[
A_1 (x_1 , B_1) := \mathbb{E}_{b_1} \left( 1\!\!1_{B_1} \Delta (x_1) \beta_1  \right), \quad B_1 \in \beta (\mathbb{R}^m),
\] where $\Delta (x) \in \mathbb{R}^{m \times p}$, for every $x \in \mathbb{R}^k$, is a feature library and $\beta_1$ has a $p$-dimensional multivariate distribution with density $g_{b_1}$ parametrized by $b_1 \in \Theta \subset \mathbb{R}^u$. We assume that $\Theta$ is a bounded convex and compact set in the Euclidean space. We denote by $D$ an upper bound on the diameter of $\Theta$:

 \[
 \parallel b_1 - b_2 \parallel \leq D, \quad \forall ~ b_ 1 , b_2 \in \Theta. 
 \] As a consequence, the action $A_1$ can be identified by the parameter $b_1 \in \Theta$. Next, we observe $y_1 \in \mathbb{R}^m$  and the learner receives the following loss

 \[
 L (b_1 , (x_1 , y_1)) = \mathbb{E}_{b_1} \left( \ell (y_1,  \Delta (x_1) \beta_1) + \lambda \beta_1^T \beta_1 \right)  
 \] where $\lambda \geq 0$ is the complexity parameter and $\ell : \mathbb{R}^m \times \mathbb{R}^m \rightarrow \mathbb{R}$ is an integrable function. 
 At the second step, we observe $x_2 \in \mathbb{R}^k$. Then the learner takes an action $A_2$ which is a probability distribution over the space of labels $\mathbb{R}^m$ as follows

\[
A_2 (x_2 , B_2) := \mathbb{E}_{b_2} \left( 1\!\!1_{B_2} \Delta (x_2) \beta_2  \right), \quad B_2 \in \beta (\mathbb{R}^m),
\] where $\beta_2$ has a $p$-dimensional multivariate distribution with density $g_{b_2}$ parametrized by $b_2 \in \Theta$. As a consequence, the action $A_2$ can be identified by the parameter $b_2 \in \Theta$. Next, we observe $y_2 \in \mathbb{R}^m$  and the learner receives the following loss

 \[
 L (b_2 , (x_2 , y_2)) = \mathbb{E}_{b_2} \left( \ell (y_2,  \Delta (x_2) \beta_2) + \lambda \beta_2^T \beta_2 \right). 
 \]
Following in this way, at the step $t$, we observe $x_t \in \mathbb{R}^k$. Then the learner takes an action $A_t$ which is a probability distribution over the space of labels $\mathbb{R}^m$ as follows

\[
A_t (x_t , B_t) := \mathbb{E}_{b_t} \left( 1\!\!1_{B_t} \Delta (x_t) \beta_t  \right), \quad B_t \in \beta (\mathbb{R}^m),
\] where $\beta_t$ has a $p$-dimensional multivariate distribution with density $g_{b_t}$ parametrized by $b_t \in \Theta$. As a consequence, the action $A_t$ can be identified by the parameter $b_t \in \Theta$. Next, we observe $y_t \in \mathbb{R}^m$  and the learner receives the following loss

 \[
 L (b_t , (x_t , y_t)) = \mathbb{E}_{b_t} \left( \ell (y_t,  \Delta (x_t) \beta_t) + \lambda \beta_t^T \beta_t \right). 
 \]  The goal of a sequential learning algorithm is to find a sequence $ \{\hat{b}^{t}: t \geq 1\}$ that achieve the minimum regret along the whole learning process. The regret is defined as

 \[
 Regret = \sum_{t=1}^n L(\hat{b}^{t} ({\bf o}_{t-1}) , (x_t,y_t)) - \min_{b \in \Theta} \sum_{t=1}^n L (b , (x_t,y_t)).
 \] In order to establish that the proximal gradient algorithm minimizes the regret, we suppose that the following hypothesis is valid.

{\bf Sequential convexity Hypothesis:}

 \begin{itemize} 

 \item For every fixed $(x,y) \in \mathbb{R}^k \times \mathbb{R}^m$, the function $L( \cdot , (x,y)): \Theta \rightarrow \mathbb{R}$ is given by

 \[
 L (b , (x,y)) = \mathbb{E}_{b} \left( \ell (y,  \Delta (x) \beta) + \lambda \beta^T \beta \right)
 \] where $\beta$ has a $p$-dimensional multivariate distribution with density $g_{b}$ parametrized by $b \in \Theta$, is a convex and differentiable function. \\

 \item $ \parallel \nabla L(b, (x,y) \parallel \leq G$ with constant $G>0$. 

 \end{itemize} 
As a consequence, we propose the following sequential proximal gradient algorithm

\[
\hat{b}^{t} ({\bf o}_{t-1}) = prox_{\eta_{t-1},\Psi} \left[\hat{b}^{t-1} ({\bf o}_{t-2}) - \eta_{t-1} \nabla L(\hat{b}^{t-1} , (x_{t-1} , y_{t-1}))\right]
\] where $\hat{b}^{1} \in \Theta$, $t \geq 2$ and

$$\prox_{\eta ,\Psi} (v) = \arg \min_z \left\{\Psi (z) + \frac{1}{2 \eta} \parallel z-v \parallel^2  \right\},$$ with

\[ \Psi(b) = \begin{cases} 
          0 &; ~ ~ ~ b \in \Theta  \\
          \infty &; ~ ~ ~ b \not \in \Theta.
       \end{cases}
\] Next, we apply usual arguments from the theory of convex learning to prove that the sequential proximal gradient minimizes the regret.

\begin{theorem}
Under the sequential convexity hypothesis, the regret function satisfies 

\[
\frac{1}{n} \left(\sum_{t=1}^n L(\hat{b}^{t} ({\bf o}_{t-1}) , (x_t,y_t)) - \min_{b \in \Theta} \sum_{t=1}^n L (b , (x_t,y_t)) \right) \leq 3 DG \frac{1}{\sqrt{n}},
\] with $\eta_t = \frac{D}{G \sqrt{t}}$.
\end{theorem}

\begin{proof}
 We consider the following stochastic Ridge regression:

\begin{equation} \label{minifunctional_sindy}
\hat{b} = \arg\min_{b \in \Theta} \E_{b} L(b, (x,y)) = \arg\min_b  \left\{L(b, (x,y)) + \Psi(b)  \right\}
\end{equation} where $\lambda \geq 0$, $L(b, (x,y)) = \E_{b} \ell (y , \Delta (x) \beta) + \lambda \E_b \beta^T \beta $ and 

\[ \Psi(b) = \begin{cases} 
          0 &; ~ ~ ~ b \in \Theta  \\
          \infty &; ~ ~ ~ b \not \in \Theta.
       \end{cases}
\]  By convexity of $L$, we have that 

\begin{equation} \label{convexity_inequality}
L(\hat{b}^{t} ({\bf o}_{t-1}) , (x_t,y_t)) \leq L (b , (x_t,y_t)) + \langle \nabla L(\hat{b}^{t} ({\bf o}_{t-1}) , (x,y)) , \hat{b}^{t} ({\bf o}_{t-1}) - b  \rangle, \quad b \in \Theta.
\end{equation} Applying the Pythagorean theorem, we have that

\begin{eqnarray*}
    \parallel \hat{b}^{t+1} ({\bf o}_{t}) - b \parallel^2 &=&
    \parallel prox_{\eta_t,\Psi} \left[\hat{b}^{t} ({\bf o}_{t-1}) - \eta_t \nabla L(\hat{b}^{t} , (x_{t} , y_{t}))\right] - b \parallel^2 \\  &\leq& \parallel \hat{b}^{t} ({\bf o}_{t-1}) - \eta_t \nabla L(\hat{b}^{t} , (x_{t} , y_{t})) - b \parallel^2 \\ &=&
    \parallel \hat{b}^{t} ({\bf o}_{t-1}) - b \parallel^2 + \eta_t^2 \parallel \nabla L(\hat{b}^{t} , (x_{t} , y_{t})) \parallel^2 - 2 \eta_t \langle \nabla L(\hat{b}^{t} , (x_{t} , y_{t})) ,  \hat{b}^{t} ({\bf o}_{t-1}) - b\rangle ,
\end{eqnarray*} for every $b \in \Theta$. Hence, we conclude that

\begin{equation} \label{inequality_inner}
2\langle \nabla L(\hat{b}^{t} , (x_{t} , y_{t})) ,  \hat{b}^{t} ({\bf o}_{t-1}) - b\rangle \leq \frac{\parallel \hat{b}^{t} ({\bf o}_{t-1}) - b \parallel^2 - \parallel \hat{b}^{t+1} ({\bf o}_{t}) - b \parallel^2}{\eta_t} + \eta_t \parallel \nabla L(\hat{b}^{t} , (x_{t} , y_{t})) \parallel^2 . 
\end{equation} As a consequence of equations \ref{convexity_inequality} and \ref{inequality_inner}, we obtain that

\begin{eqnarray*}
L(\hat{b}^{t} ({\bf o}_{t-1}) , (x_t,y_t)) - L (b , (x_t,y_t)) &\leq& \frac{\parallel \hat{b}^{t} ({\bf o}_{t-1}) - b \parallel^2 - \parallel \hat{b}^{t+1} ({\bf o}_{t}) - b \parallel^2}{\eta_t} \\ &+& \eta_t \parallel \nabla L(\hat{b}^{t} , (x_{t} , y_{t})) \parallel^2 .
\end{eqnarray*} Summing from $1$ to $n$ with $\eta_t = \frac{D}{G \sqrt{t}}$ and $\frac{1}{\eta_0} = 0$, we obtain for every $b \in \Theta$ that

\begin{eqnarray*}
 \sum_{t=1}^n \left(L(\hat{b}^{t} ({\bf o}_{t-1}) , (x_t,y_t)) - L (b , (x_t,y_t)) \right) &\leq& \sum_{t=1}^n \parallel \hat{b}^{t} ({\bf o}_{t-1}) - b \parallel^2 \left( \frac{1}{\eta_t} - \frac{1}{\eta_{t-1}} \right) \\ &+&  \sum_{t=1}^n \eta_t \parallel \nabla L(\hat{b}^{t} , (x_{t} , y_{t})) \parallel^2 \\ &\leq& D^2 \sum_{t=1}^n \left( \frac{1}{\eta_t} - \frac{1}{\eta_{t-1}} \right) + G^2 \sum_{t=1}^n \eta_t \\ &\leq& D^2 \frac{1}{\eta_n} + DG \sum_{t=1}^n \frac{1}{\sqrt{t}} \leq DG\sqrt{n} + DG 2\sqrt{n} \leq 3 DG\sqrt{n} .
\end{eqnarray*} The last inequality follows since $\sum_{t=1}^n \frac{1}{\sqrt{t}} \leq 2 \sqrt{n}$.

\end{proof}
If we consider the $p$-dimensional vector $\hat{\beta}_t$ with distribution $g_{\hat{b}_t}$, we can predict the output of the process as follows

\[
\hat{y}_{t} = \mathbb{E}_{\hat{b}_t} \left( \Delta (x_{t}) \hat{\beta}_t  \right), \quad t \geq 1.
\] As we know the distribution of the quantity $\Delta (x_{t}) \hat{\beta}_t $, the prediction probability interval is given by 

\[
\mathbb{P} \left[ LI_{\alpha/2} \leq \Delta (x_{t}) \hat{\beta}_t \leq  LS_{\alpha/2} \right] = 1 - \alpha, \quad 0 < \alpha < 1.
\]

\subsection{Randomized SINDy for regression}
Considering a sequential regression model $y= \Delta(x) \beta$ where $\beta = \left(\beta_1, \cdots, \beta_p \right)$.  If we assume $\beta$ follows a multivariate normal distribution $\mathcal{N}_p(\mu, \Sigma)$ with mean vector $\mu = (\mu_{1} , \cdots , \mu_{p})$, covariance matrix $\Sigma = \diag (\sigma_{1} , \cdots , \sigma_{p})$ and density function given by:
$$ 
g_b(\beta) = \frac{1}{(2\pi)^{p/2} |\Sigma|^{1/2}} \exp \left\{-\frac{1}{2}  
(\beta - \mu)^\top \Sigma^{-1} 
(\beta - \mu) \right\}.
$$

In this case, we have $b = \left(\mu,\Sigma \right)$, where $\mu \in \mathbb{R}^p$ and $\Sigma \in \mathbb{R}^{p\times p}$ (p.s.d.).  By assuming this parametric form for $g$, the constraints for a probability density are inherently satisfied. At every time step, the density function $g_b(\beta)$ is completely determined by $b$. By using the quadratic loss, the optimization problem, according to \autoref{minifunctional_sindy}, is as follows:
\begin{equation} \label{opt}
    \argmin_{\mu, \Sigma} \left\{~ 
    \E~(y - \Delta(x) \beta)^2 + 
    \lambda \E~ \beta^\top \beta + \Psi(\Sigma) 
 ~\right\}
\end{equation} where $\lambda \geq 0$ and 
\[ \Psi(\Sigma) = \begin{cases} 
          0 &; ~ \text{if }\Sigma\text{ is positive semi-definite (p.s.d.),}   \\
          \infty &; ~\text{otherwise}.
       \end{cases}
\]

We now compute each term in the objective function \ref{opt}. Let $z:= \Delta(x)$ and since $\beta_p \sim \mathcal{N}(\mu, \Sigma)$, 
$$(y -  \zt \beta)^2 = y^2 - 2y  \zt \beta + ( \zt \beta)^2,$$
the expected quadratic loss becomes
\begin{eqnarray*}
     \E (y -  \zt \beta)^2 &=& \E y^2 - 2y \zt \E \beta + \E (\zt \beta)^2.
\end{eqnarray*}
We know that
$$\E[\beta] = \mu,$$
$$\text{Var}(\beta) = \Sigma = \E[\beta\beta^\top] - \E[\beta]\E[\beta]^\top \implies \E[\beta\beta^\top]  = \Sigma + \mu \mu^\top .$$
Thus,
\begin{eqnarray*}
\E~(\zt \beta)^2 &=& \E~\zt \beta \beta^\top z \\
&=& \zt \E[\beta\beta^\top] z\\
&=& \zt (\Sigma + \mu \mu^\top) z \\
&=& \zt \Sigma z + \zt \mu \mu^\top z \\
&=& \zt \Sigma z + (\zt \mu)^2.
\end{eqnarray*}
Substituting these into the expected quadratic loss, we obtain
$$
\E \left[ (y -  \zt \beta)^2\right]
= y^2 - 2y \zt \mu + \zt \Sigma  z + ( \zt \mu)^2.
$$

For $\lambda > 0$, the expectation of $L_2$ regularization term is simply
$$
\lambda \E \beta^\top \beta = \lambda \left( \text{tr} \Sigma + \mu^\top \mu \right). 
$$

And the constraint term enforces that the estimated covariance matrix of $\beta$ is p.s.d. which can be represented using an indicator function:
$$
1\!\!1_{p.s.d}: \Sigma \rightarrow \{0,1\}.
$$

We can apply proximal gradient descent to minimize the objective function which can be treated as two parts, $f_1$ and $f_2$:
$$
\underbrace{
    y^2 - 2y \zt \mu + \zt \Sigma  \mathbf{x} + ( \zt \mu)^2 +  \lambda \left( \text{tr} \Sigma + \mu^\top \mu \right)
}_{f_1} + 
\underbrace{
    1\!\!1_{p.s.d}(\Sigma)
}_{f_2}
$$

The partial derivative with respect to $\mu$ is:
$$ \nabla_{\mu} f_1 (\mu, \Sigma) =
2z (z^\top \mu - y) + 2\lambda\mu,$$
and the partial derivative with respect to the $\Sigma$ is:
$$ \nabla_{\Sigma} f_1(\mu,\Sigma) = 
z z^\top + \lambda I. $$

The indicator function in $f_2$ is non-smooth and the proximal operator is reduced to a projection onto a p.s.d. cone, following \cite{CVX}, 
$$
    \text{proj}_{K}(\Sigma) = \sum_{i=1}^p \max\{0,\zeta_i\}v_iv_i^\top,
$$
where $K$ is the p.s.d. cone and $\Sigma = \sum_{i=1}^p \zeta_i v_iv_i^\top$ is the eigenvalue decomposition of $\Sigma$.

Therefore, at the end of every time step, we update the parameters of $g_{t+1}$ based on the observed data $(z_t, y_t)$:
\begin{equation}
\label{eq: iter_mu}
\mu^{t+1} = \mu^t - \eta_t \nabla_{\mu^t} f_1 (\mu^t, \Sigma^t)  = \mu^t - \eta_t \left( 2 z_t \left(z_t^\top \mu^t - y_t \right) + 2 \lambda \mu^t 
\right),
\end{equation}

\begin{equation*}
\widetilde{\Sigma}^{t+1} = \Sigma^t - \eta_t \nabla_{\Sigma^t} f_1(\mu^t,\Sigma^t)  = \Sigma^t - \eta_t \left( z_t \zt_t + \lambda I \right),
\end{equation*}

\begin{equation}
\label{eq: iter_Sigma}
\Sigma^{t+1} = \text{proj}_K \left(  \widetilde{\Sigma}^{t+1} \right)
\end{equation}
where $\eta_t > 0$, $\lambda \geq 0$.

The prediction for the next output at step $t+1$, given the observed feature vector $z_{t+1}$, is computed as
$$
\hat{y}_{t+1} = z_{t+1}^\top \hat{\beta}_t, \text{ where } \hat{\beta}_t=\mu^t
$$
and the corresponding $100(1-\alpha)\%$ prediction probability interval is given by
$$
\hat{y}_{t+1} \pm  z_{\alpha/2} \sqrt{\widehat{Var}(\beta)_t}, \text{ where } \widehat{Var}(\beta)_t =  z_{t+1}^\top \Sigma^t z_{t+1} \text{ and } 0<\alpha <1.
$$

\subsection{Randomized SINDy for binary outcome}
In a binary classification setting where $Y \in \{0,1\}$ and $z := \Delta(x) \in \mathbb{R}^p$, we can adapt the previous example by utilizing the logistic function such that $h(z) := \sigma(\zt \beta)$ represents the probability of predicting the output label $Y=1$:
$$
P(Y=1 \mid z, \beta) = \sigma(\zt \beta) = \dfrac{1}{1 + \exp\{-\zt \beta\}}.
$$

Using the logistic loss function,
$$
\ell(y, \zt \beta) = -\bigr[ y \ln \left(h\left(z\right)\right)+\left(1-y\right) \ln \left(1-h\left(z\right)\right) \bigr],$$
we note that there is no closed form for $\E \ell(y, \zt \beta)$. Rather, we consider its approximation
$\ell (y, \zt \mu)$. Consequently, the objective function is as follows:
$$
\underbrace{\ell(y, \zt \mu) + 
\lambda \left( \text{tr} \Sigma + \mu^\top \mu \right)}_{f_1}  + \underbrace{
    1\!\!1_{p.s.d}(\Sigma)
}_{f_2} 
$$

The partial derivative with respect to $\mu$ is:
$$
\nabla_{\mu} f_1 (\mu, \Sigma) = 
z \left(\sigma \left(\zt\mu \right) - y \right) + 2\lambda\mu,
$$
and the partial derivative with respect to the $\Sigma$ is:
$$ \nabla_{\Sigma} f_1(\mu,\Sigma) = 
\lambda I
$$

Therefore, at the end of every time step, we update the parameters of $g_t$ based on the observed data $(\mathbf{x}_t, y_t)$:
$$ \mu_{t+1} = \mu_t - \eta_t \nabla_{\mu_t} f_1 (\mu_t, \Sigma_t)  = \mu_t - \eta_t \left( \mathbf{x}_t \left( \frac{1}{1 + e^{-\zt_t \mu_t}} - y_t \right) + 2 \lambda \mu_t 
\right),$$

$$ \widetilde{\Sigma}_{t+1} = \Sigma_t - \eta_t \nabla_{\Sigma_t} f_1(\mu_t,\Sigma_t)  = \Sigma_t - \eta_t \left( \lambda I \right),$$

$$
\Sigma_{t+1} = \text{proj}_K \left(  \widetilde{\Sigma}_{t+1} \right)
$$
where $\eta_t > 0$, $\lambda \geq 0$.


\section{Experiments} \label{sec:experiments}

In this section, we present some simulation results of models discussed in \autoref{sec:RSindy}. All experiments were executed using the R language.

\subsection*{Experiment 1} 
\begin{figure}[ht]
    \centering
    \includegraphics[scale=.4]{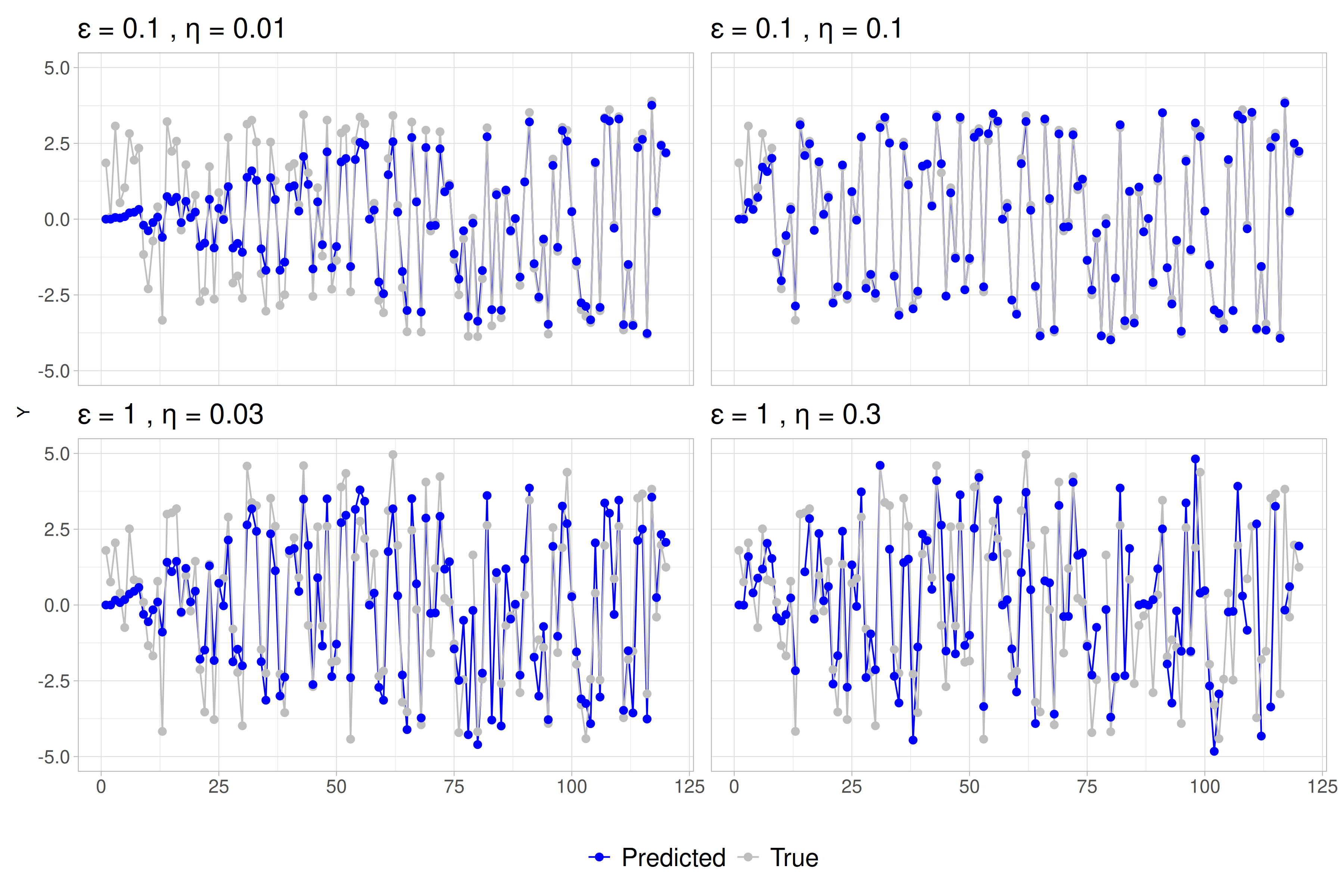}
    \caption{Study of the impact of noise on the selection of learning rate}
    \label{fig:ex4.2_1}
\end{figure}

Considering an outcome sequence $Y = 2X + \varepsilon$ where $\varepsilon \sim N(0, \epsilon)$, the data is generated by selecting $X \sim U(-2,2)$. To analyze the impact of noise and the choice of learning rate $\eta$, we vary the noise level between 0.1 and 1. For the low-noise case ($\epsilon = 0.1$), we consider learning rates of 0.01 and 0.1. In the high-noise case ($\epsilon=1$), we vary $\eta$ between 0.03 and 0.3. In both cases, the initial value for parameter is set to $\mu_0 = 0$, also, no regularized term is considered ($\lambda = 0$). We simulated a sequence of size 120 and the model quality results are summarized in \autoref{tab:ex4.2_1}. Additionally, \autoref{fig:ex4.2_1} presents a comparison between the true values of $y$ and the predicted values across different scenarios.

\begin{table}[ht]
\caption{Simulation results} \label{tab:ex4.2_1}
\begin{tabular}{c|cccc}
\toprule
$\epsilon$ & $\eta$ & $R^2 $ & $\hat{\sigma}$ & RSME   \\
\midrule
\multirow{2}{*}{0.1}    & 0.01 & 0.8490 & 0,9305 & 0,9267 \\
                        & 0.1  & 0.9805 & 0.3341 & 0.3327 \\
\midrule
\multirow{2}{*}{1}      & 0.03 & 0.7973 & 1.1933 & 1.188 \\
                        & 0.3  & 0.4156 & 2.0260 & 2.0175 \\
\bottomrule
\end{tabular}
\end{table}

For low-noise data, a higher learning rate achieves a better R-squared value and an estimated $\sigma$ closer to the true noise level. However, in high-noise scenarios, a lower learning rate is preferable, as it helps stabilize learning and prevent erratic fluctuations in parameter updates. As observed in the graph, higher learning rate results in larger prediction errors. This suggests that a higher learning rate prioritizes adjustments based on recent data, while a lower learning rate means to retain information given by older data entries. We conducted a Monte Carlo simulation with 1,000 replicates, each consisting of a sequence size of 10,000 observations. The simulation results are shown in \autoref{tab:ex4.2_2}.

\begin{table}[ht]
\caption{Simulation results}
\label{tab:ex4.2_2}
\begin{tabular}{c|ccccc}
\toprule
$\epsilon$ & $\eta$ & $R^2$     & $\hat{\sigma}$ & RSME   & $\hat{\beta}$  \\
\midrule
0.1 & 0.1  & 0.9976 & 0.1144 & 0.1144  & 2.000 (1.9979, 2.002)\\
1   & 0.03 & 0.8346 & 1.0232 & 1.0232 & 2.004 (1.9893, 2.0107)\\
\bottomrule
\end{tabular}
\end{table}

\begin{figure}[ht]
    \centering
    \includegraphics[scale=.45]{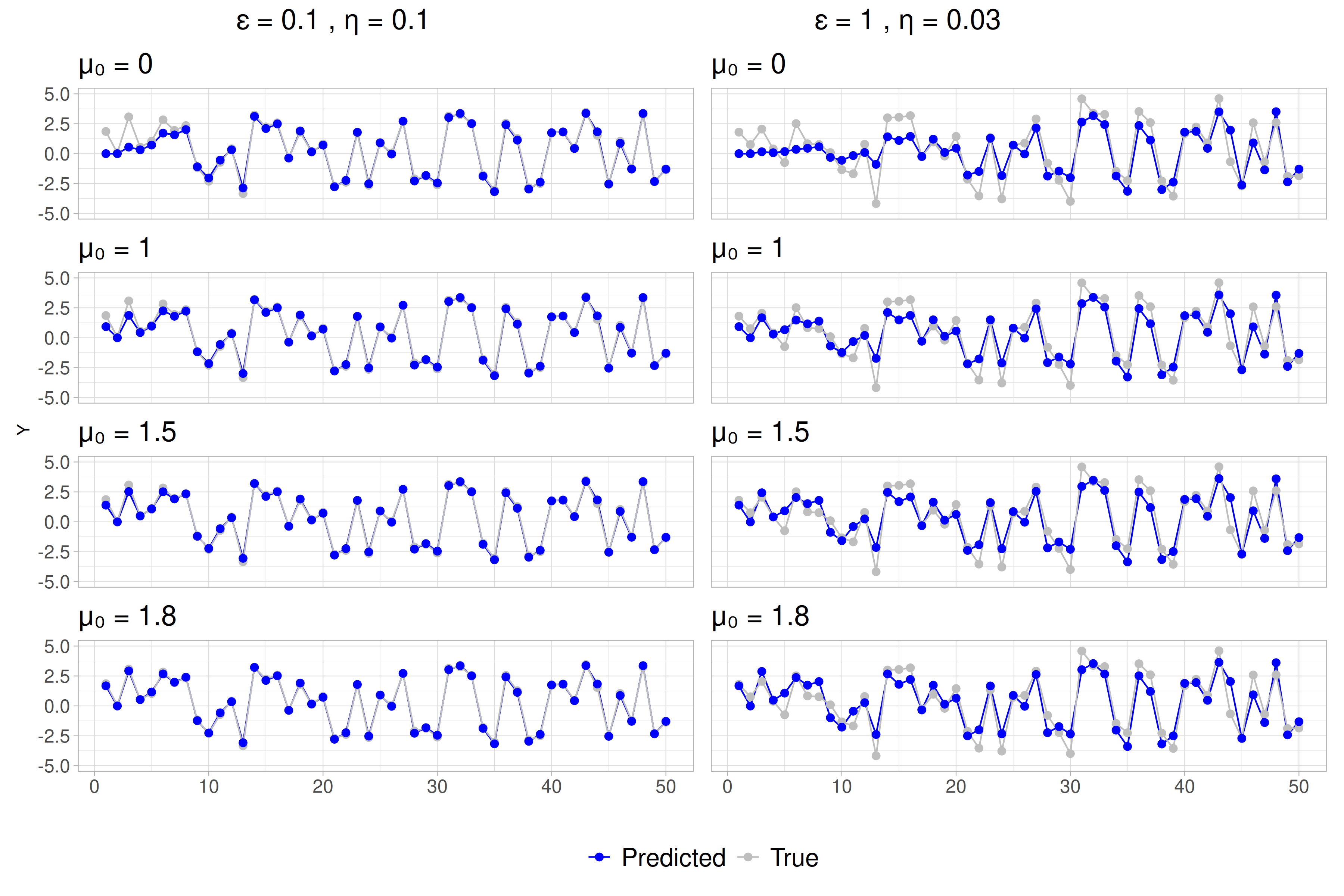}
    \caption{Study of the impact of initial values on convergence}
    \label{fig:ex4.2_2}
\end{figure}

To evaluate the impact of the initial value, we vary $\mu_0$ between 0 and 1.8. As shown in \autoref{fig:ex4.2_2}, the initial values influence the convergence rate. For a sequence observed at the 50th time step, the initial value of $\mu_0=0$ requires more iterations to achieve convergence.

\subsection*{Experiment 2}
We created an ill-conditioned model matrix for simulation purpose and the predictors were generated as follows: $X_1 \sim \mathcal{N}(0,1)$, $X_2 = X_1 + \delta_2$, $X_3 =2X_1 + 3X_2 + \delta_3$, where $\delta_2 \sim N(0,0.01^2)$, $\delta_3 \sim N(0,0.01^2)$. The outcome sequence was defined as $Y=-1+2X_1-2X_2 + 1.5X_3 + \varepsilon$, with $\varepsilon \sim N(0,1)$.   
A sequence of 500 time steps was generated. 

The first 250 time steps were designated as a warm-up period for model estimation. Within this warm-up period, we performed a rolling window cross-validation with 40 train-validation splits. The first training set consisted of 50 observations. For each subsequent split, the training set size was incremented by five observations (e.g.. 50, 55, ..., up to 245 observations). Each corresponding validation set contained the five subsequent observations immediately following its training set. For each of these 40 train-validation splits, we fitted a regularized regression model. The regularization parameter ($\lambda$) was varied across 30 different values, ranging from 0 to 1. To determine the optimal $\lambda$, the minimum Root Mean Squared Error (RMSE) was calculated for each $\lambda$ value on each of the 40 validation sets. Subsequently, $\lambda = 0.3448$ was selected as the optimal $\lambda$ as it yielded the lowest overall RMSE.  
Still within this warm-up period, we proceeded to standardize the model matrix, which has dimensions of $250 \times 4$. A preliminary Tikhonov regularization was then performed to obtain an initial guess for the mean coefficient vector ($\mu_0$) and its covariance matrix ($\Sigma_0)$. Subsequently, the algorithm was applied with the parameter update steps defined by Equations \ref{eq: iter_mu} and \ref{eq: iter_Sigma}. The model parameters estimated at $t=250$ were used as the initial state for the subsequent time step. From $t=251$ onwards, each new incoming data dynamically updated both the standardization of the model matrix and the estimated parameters. With a learning rate of $\eta=1 \times 10^{-3}$, \autoref{tab:sim2} presents the estimated parameters at the initial and the final time step.

\begin{table}[ht]
    \centering
    \caption{Estimated parameters at initial and final iterative step of the warm-up estimation period}
    \label{tab:sim2}
    \begin{tabular}{c|ccccc}
    \toprule
    Parameter & $t$ & intercept & $X_1$ & $X_2$ & $X_3$ \\
    \midrule     
    \multirow{2}{*}{$\hat{\mu}_t$} & 1    & -1.5652 & 2.5442 & 2.2182 & 2.2142\\
    & 500  & -1.3349 & 2.2946 & 2.0639 & 2.0610\\
    \midrule
    \multirow{2}{*}{$\widehat{\Sigma}_t$ } & 1   & 0.0075 & 0.1271 & 0.0952 & 0.0112 \\
 & 500 & 0.0000 & 0.0267 & 0.0195 & 0.0006 \\
    \bottomrule
    \end{tabular}
\end{table}

Model quality metrics at the final time step are presented in \autoref{tab:sim2_quality}:
\begin{table}[ht]
    \caption{Model quality measure at the final time step}
    \label{tab:sim2_quality}
    \centering
    \begin{tabular}{cccccccc}
    \toprule
    $t$ & $SST$ & $SSE$ & $n$ & $p$ & $R^2$ & $\hat{\sigma}$ & $RMSE$\\
    \midrule
 500 & 25216.07 & 738.6398 & 500 & 4 & 0.97071 & 1.22032 & 1.21543\\
    \bottomrule
    \end{tabular}
\end{table}


\subsection*{Experiment 3}
We simulated a binary sequence $Y \sim Ber(p)$, where $p$ was defined by the logistic model $P(Y=1 \mid X) = \dfrac{1}{1 + \exp\{-(1+2X_1 + 3X_2)\}}.$
As in the previous experiment, a sequence of 500 time steps was generated. Initial logistic model parameters were obtained by fitting a ridge logistic regression with $\lambda = 0$. The algorithm from Example 2 of \autoref{sec:RSindy} was then applied with a fixed learning rate of $\eta=0.1$ . \autoref{tab:sim3} shows the estimated coefficients at both the initial and final time steps.
\begin{table}[ht]
    \centering
    \caption{Estimated parameters at initial and final iterative step of the warm-up estimation period}
    \label{tab:sim3}
    \begin{tabular}{cccc}
    \toprule
    $t$ & intercept & $X_1$ & $X_2$\\
    \midrule     
    1    & 0.7897 & 1.9020 & 2.8405\\
    500  & 0.8002 & 1.9081 & 2.8414\\
    \bottomrule
    \end{tabular}
\end{table}

\begin{figure}
    \centering
    \includegraphics[width=0.4\linewidth]{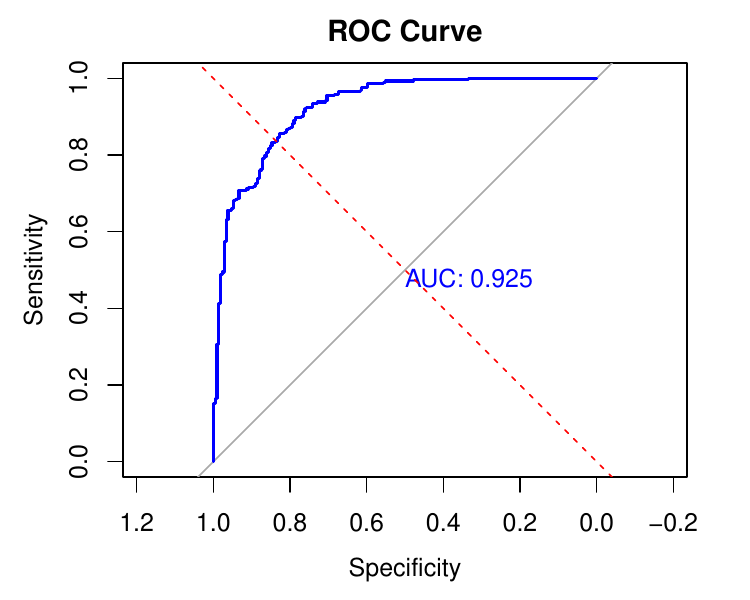}
    \caption{ROC curve}
    \label{fig:sim3_roc}
\end{figure}
The Receiver Operating Characteristic (ROC) curve, \autoref{fig:sim3_roc}, indicated an optimal threshold of 0.5745, resulting in an Area Under the Curve (AUC) of 0.925. Additional model quality metrics, such as an accuracy greater than 80\% (\autoref{tab:sim3_quality}), confirm the model's strong performance.
\begin{table}[ht]
    \caption{Model quality measure}
    \label{tab:sim3_quality}
    \centering
    \begin{tabular}{cccccccc}
    \toprule
    Accuracy & TPR & TNR & Precision & F1 & Entropy\\
    \midrule
 0.842 & 0.8522 & 0.8278 & 0.8732 & 0.8626 & 2.5467\\
    \bottomrule
    \end{tabular}
\end{table}

\subsection*{Experiment 4}

We applied the models discussed in \autoref{sec:RSindy} to forecast the monthly U.S. unemployment rate. As described in the work of \cite{Elec2}, the independent variables included are Initial Claims (IC) for unemployment insurance, the inflation rate (measured as the seasonal growth rate of the Consumer Price Index), and industrial
growth (the seasonal growth rate of the Industrial Production Index). All data were sourced from the FRED database (\url{https://fred.stlouisfed.org/}). The analysis period spans from January 1967 to May 2025. \autoref{tab:UNRATE_vars} gives a brief description of the variables used for forecasting. While most variables are originally monthly, the weekly Initial Claims data was averaged to a monthly frequency.

\begin{table}[ht]
    \centering
\caption{Variable descriptions for U.S. unemployment rate forecasting, data covering January 1967 to May 2025}
\label{tab:UNRATE_vars}
    \begin{tabular}{c|c}
    \toprule
    Variable&  Description\\
    \midrule
         UNEMP&  The monthly unemployment rate (as a percentage)\\
         UNEMP\_LAG1 &  The monthly unemployment rate (as a percentage) from previous month\\
         IC    &  The number of Initial Claims (IC) of unemployment insurance \\
         CPI   &  The Consumer Price Index (CPI)\\
         IPI & The Industrial Production Index (IPI) \\
    \bottomrule
    \end{tabular}
\end{table}

Since the unemployment rate is non-negative, we applied a transformation to the response variable $Y = \log \left( \frac{UNEMP}{100-UNEMP}\right)$. Based on the Akaike Information Criterion (AIC), the inclusion of the variables in \autoref{tab:UNRATE_vars} is validated using the \texttt{auto.arima} function from the \texttt{forecast} package. The model that included the lagged unemployment rate and three other explanatory variables obtained the smallest AIC as shown in \autoref{tab:UNRATE_AIC}.

\begin{table}[ht]
    \centering
\caption{AIC comparison for forecasting models}
\label{tab:UNRATE_AIC}
    \begin{tabular}{c|c}
    \toprule
    Model &  AIC\\
    \midrule      
         UNEMP\_ LAG1 &  -1661.203 \\
         UNEMP\_ LAG1 \& IC &  -1700.652 \\
         UNEMP\_ LAG1, IC \& CPI   &  -1761.693\\
         UNEMP\_ LAG1, IC, CPI \& IPI & -1772.673 \\
    \bottomrule
    \end{tabular}
\end{table}

\begin{figure}
    \centering
    \begin{subfigure}{0.45\linewidth}
    \includegraphics[width=\linewidth]{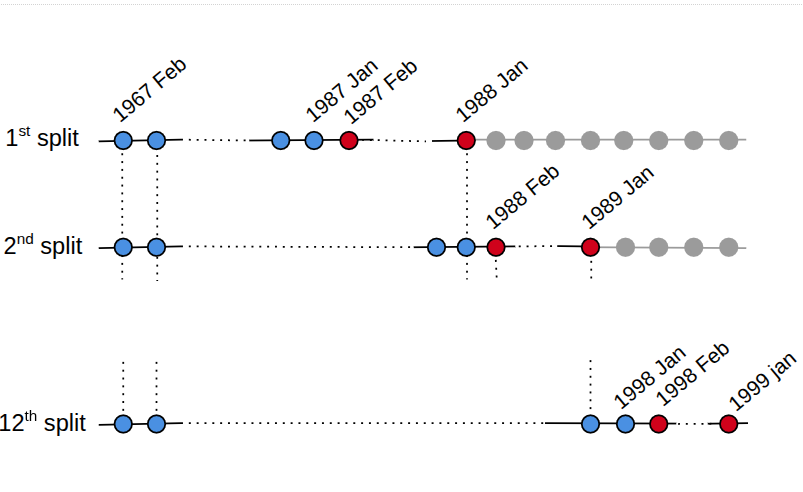}
    \end{subfigure}
    \begin{subfigure}{0.45\linewidth}
    \includegraphics[width=\linewidth]{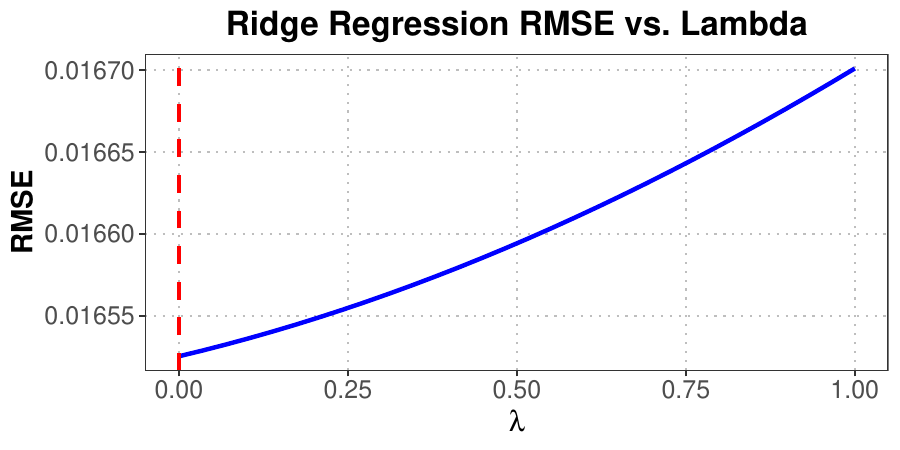}
    \end{subfigure}
    \caption{Train-validation set splits for hyperparameter selection}
    \label{fig:UNRATE_best_lambda}
\end{figure}
We used the initial 30 years of data, spanning from January 1967 to December 1999, as a warm-up period for our online learning model. Based on a rolling expanding window approach, we began with 20 years of data for training to make a one-year-ahead prediction. This process generated 12 distinct train-validation set splits for model evaluation. For each of these 12 training sets, we performed regression with Tikhonov regularization, exploring 30 different regularization parameter ($\lambda$) values ranging from 0 to 1. The optimal regularization term was identified as being the value of 0, as this produced the minimum overall RMSE (see \autoref{fig:UNRATE_best_lambda}). We then proceeded to standardize the model matrix, which had dimensions of 395 × 5. A preliminary Tikhonov regularization was then performed to obtain an initial guess for the mean coefficient vector ($\mu_0$) and its covariance matrix ($\Sigma_0$).The algorithm was then applied with the parameter update steps defined by Equations \ref{eq: iter_mu} and \ref{eq: iter_Sigma}. \autoref{tab:UNRATE_params1} presents the estimated parameters from both the initial guess and the final iterative step of the warm-up estimation period.

\begin{table}[ht]
    \centering
    \caption{Estimated parameters at initial and final iterative step of the warm-up estimation period}
    \label{tab:UNRATE_params1}
    \begin{tabular}{c|cccccc}
    \toprule
    Parameter & $t$ & intercept & UNEMP\_LAG1 & IC & CPI & IPI \\
    \midrule
     
    \multirow{2}{*}{$\hat{\mu}_t$}          & 1       & -2.7536 & 0.2461 & 0.0272 & 0.0054 & -0.0132\\
                                            & 395     & -2.7533 & 0.2454 & 0.0267 & 0.0051 & -0.0137\\
    \midrule
    \multirow{2}{*}{$\widehat{\Sigma}_t$ }  & 1       & 0.0064 & 0.0227 & 0.0204 & 0.0986 &  0.0968\\
                                            & 395     & 0.0051 & 0.0079 & 0.0086 & 0.0971 &  0.0835\\
    \bottomrule
    \end{tabular}
\end{table}

\begin{figure}[ht]
    \centering
    \includegraphics[width=\linewidth]{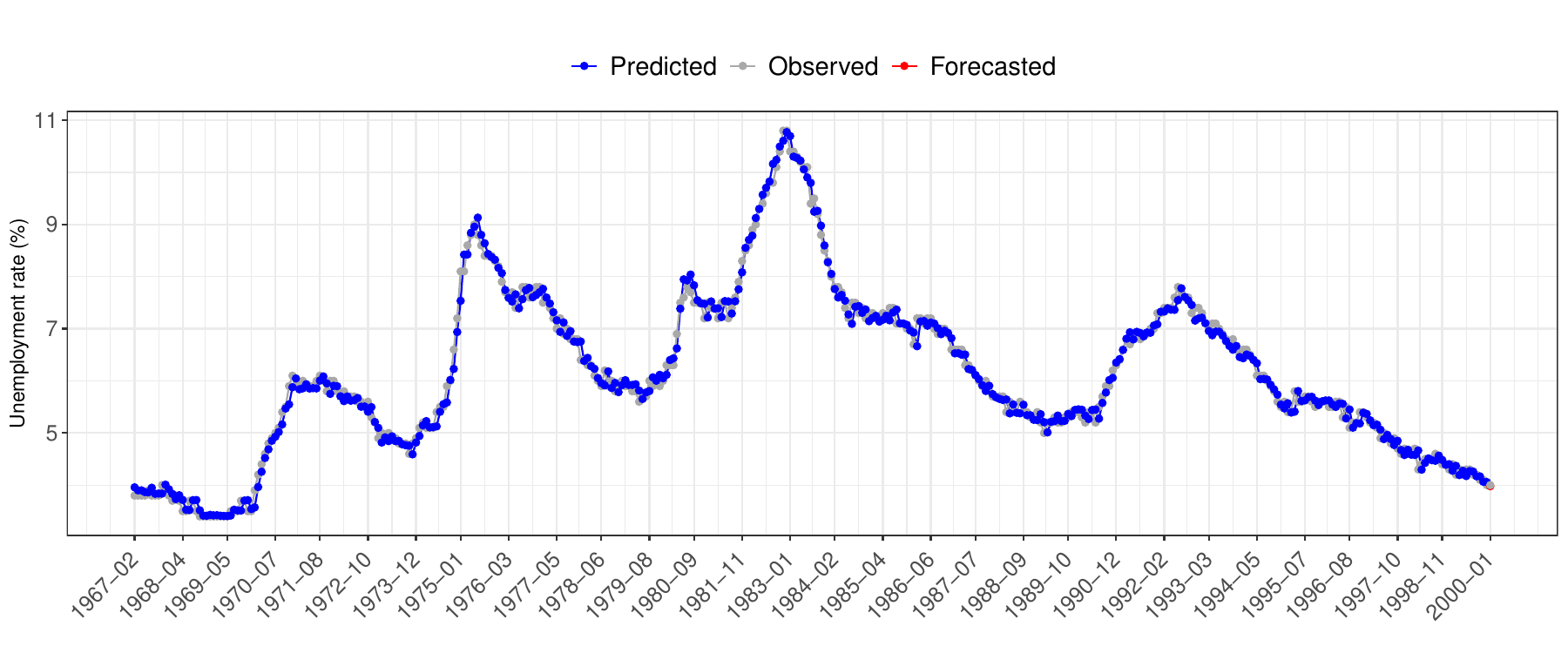}
    \caption{U.S. unemployment rate spanning from January 1967 to January 2000. Predicted values (blue), actual values (gray) and January 2000 forecast (red dot)}
    \label{fig:UNRATE_warmup}
\end{figure}

\begin{figure}{
    \centering
    \raisebox{-\height}{%
        \begin{subfigure}{0.45\linewidth}
            \includegraphics[width=\linewidth]{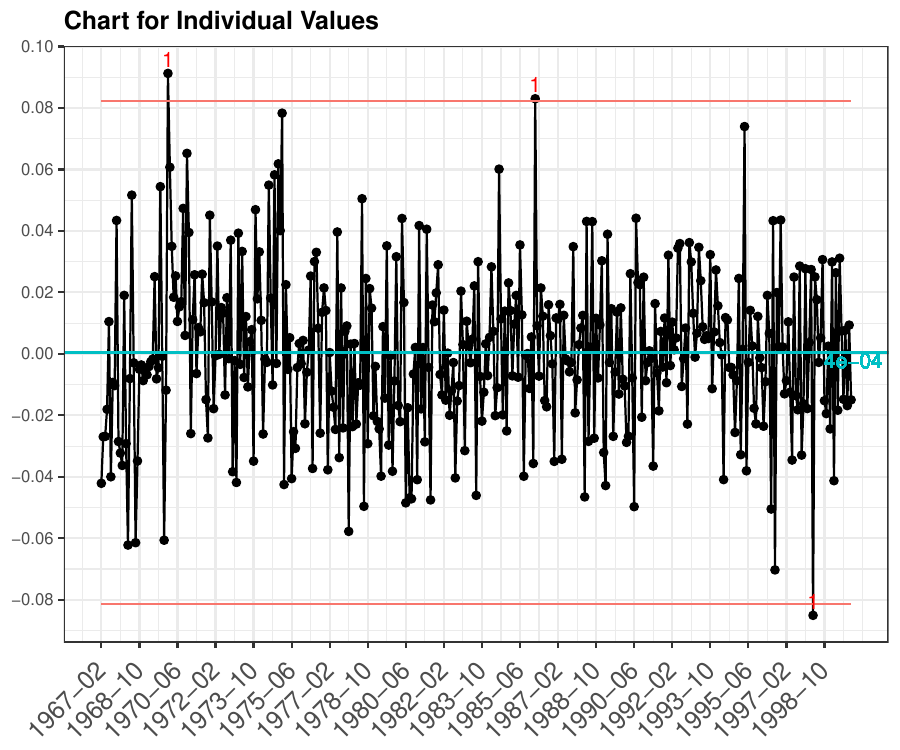}
        \end{subfigure}
        }
    \raisebox{-2.5\height}{
        \begin{subfigure}{0.45\linewidth}
            \begin{tabular}{p{4cm} c}
                \toprule
                 &  Values\\ \midrule
                Upper Limit  & 0.0821 \\
                Central Line & 0.0004 \\
                Lower Limit  & -0.0813 \\
                \bottomrule
            \end{tabular}
        \end{subfigure}
    }
    \subcaption{Individual residual values chart}
    \label{fig:UNARTE_warmup_res1}
    \begin{subfigure}{0.45\linewidth}
    \includegraphics[width=\linewidth]
    {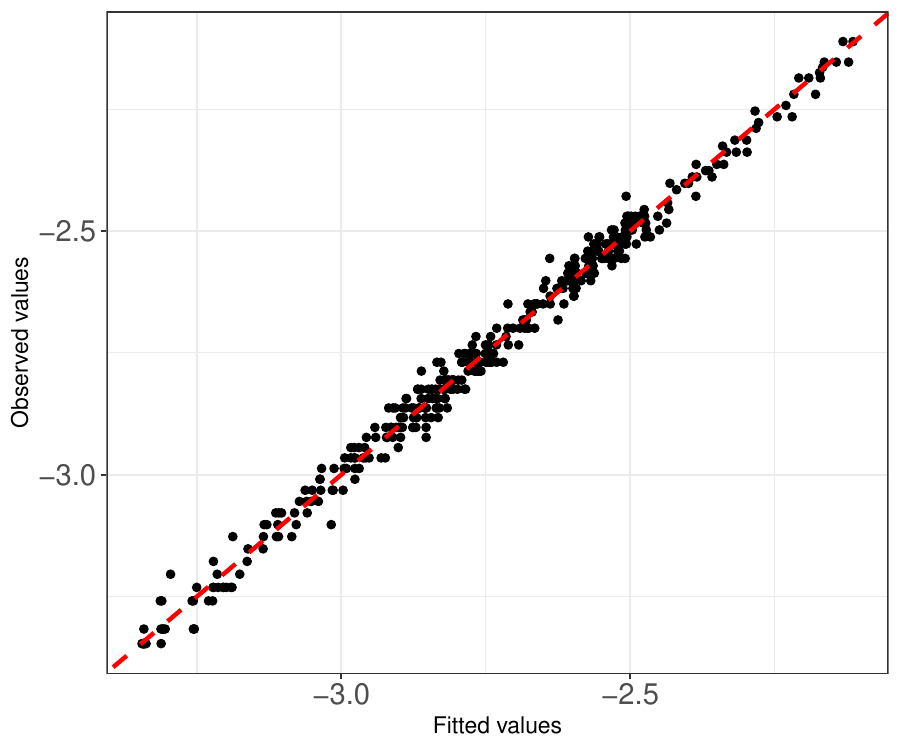}
    \caption{Observed v.s. fitted values}
    \end{subfigure}
    \begin{subfigure}{0.45\linewidth}
        \includegraphics[width=\linewidth]{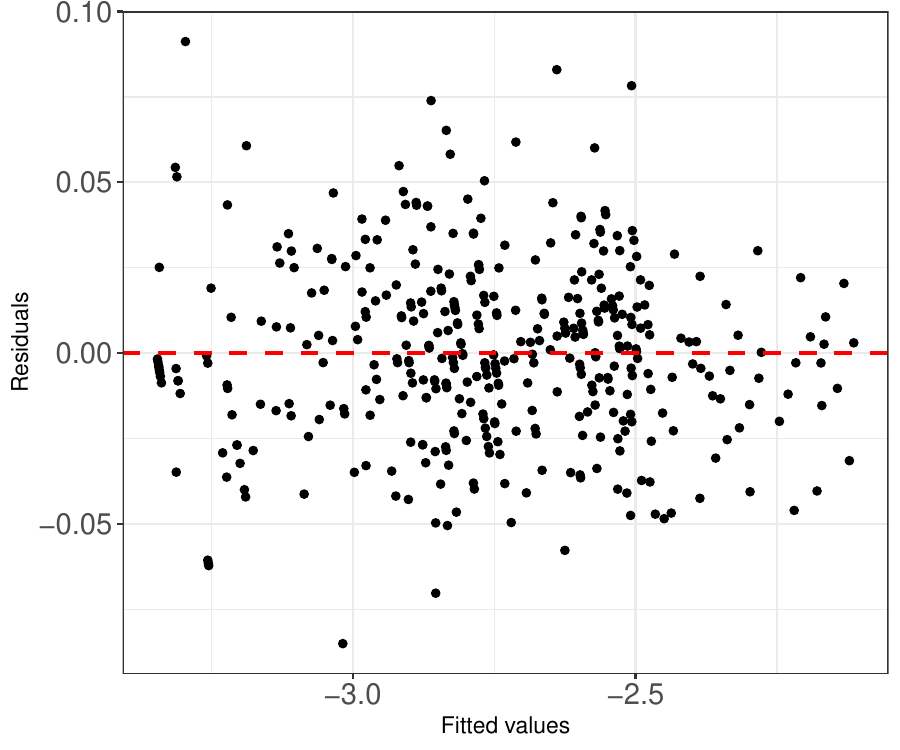}
        \caption{Residuals plot}
        \label{fig:UNARTE_warmup_res4}
    \end{subfigure}
    \raisebox{-\height}{%
        \begin{subfigure}{0.45\linewidth}
            \includegraphics[width=\linewidth]{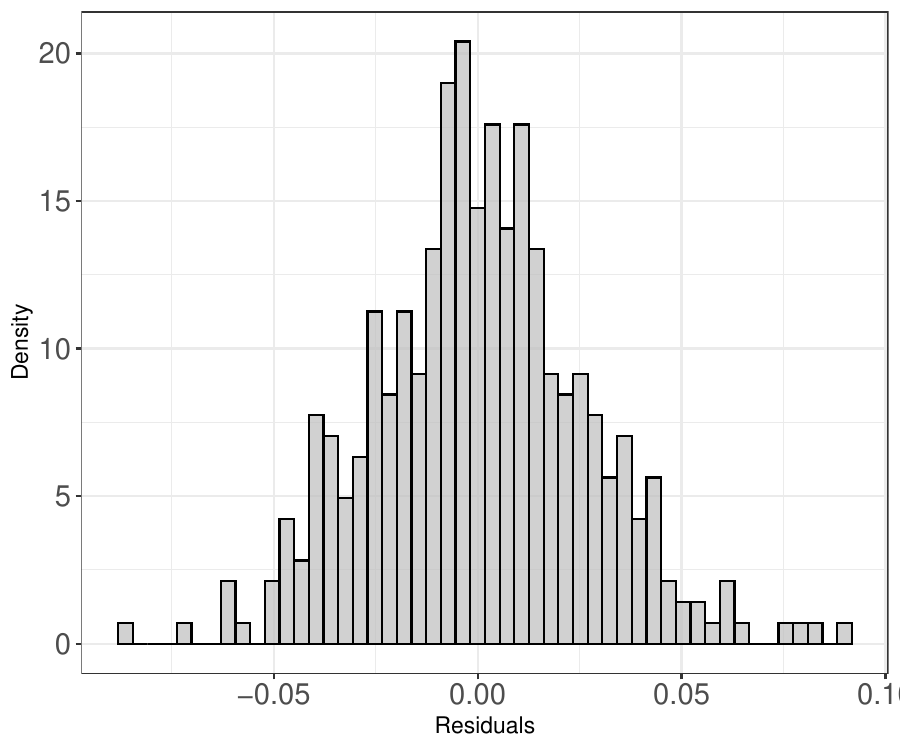}
        \end{subfigure}
    }
    \raisebox{-1.5\height}{
        \begin{subfigure}{0.45\linewidth}
            \begin{tabular}{p{4cm} c}
                \toprule
                 &  Values\\ \midrule
                Min.    & -0.0851 \\  
                1st Qu. & -0.0165 \\  
                Median  & -0.0006 \\  
                Mean    & 0.0004 \\  
                3rd Qu. & 0.0159 \\  
                Max.    & 0.0912 \\ 
                \bottomrule
            \end{tabular}
           \end{subfigure}
    }
    \subcaption{Exploratory of residuals}
    \label{fig:UNARTE_warmup_res5}
    }
    \caption{Residuals analysis for the warm-up estimation period}
    \label{fig:UNRATE_warmup_residuals}
\end{figure}

\autoref{fig:UNRATE_warmup} illustrates the predicted employment rates along with the true employment rates on the original scale. The close alignment between the predicted and actual values is evident, with an $R^2$ of 0.99 indicating a very strong fit (\autoref{tab:UNRATE_quality1}). Using the estimated parameters from the final warm-up period time step ($t=395$), we made a forecast for the following month's unemployment rate. \autoref{tab:UNRATE_forecast1} presents this forecasted value with its 95\% uncertainty level, which can be compared to the true observed value in both transformed and original scale (in brackets). The results of the residual analysis are presented in \autoref{fig:UNRATE_warmup_residuals}.

\begin{table}
    \caption{Model quality measure at the final time step in warm-up estimation period}
    \label{tab:UNRATE_quality1}
    \centering
    \begin{tabular}{cccccccc}
    \toprule
    Date & $SST$ & $SSE$ & $n$ & $p$ & $R^2$ & $\hat{\sigma}$ & $RMSE$\\
    \midrule
    1999-12 & 29.1597 & 0.2716 & 395 & 5 & 0.9907 & 0.0264 & 0.0262\\
    \bottomrule
    \end{tabular}
\end{table}

\begin{table}
    \caption{Forecasted value and prediction interval}
    \label{tab:UNRATE_forecast1}
    \centering
    \begin{tabular}{ccccc}
      \toprule
      Date  & Forecasted & Lower limit & Upper limit & True value\\
      \midrule
    2000-01   & -3.1836 & -3.3108 & -3.0563 & -3.1781 \\
              &  (3.9789) &  (3.5201) &  (4.4947) & (4) \\
    \bottomrule
    \end{tabular}
\end{table}

The Individual values control chart in \autoref{fig:UNARTE_warmup_res1} confirms that the residuals are randomly distributed with a zero mean. This interpretation is consistent with the residuals plot in \autoref{fig:UNARTE_warmup_res4}. Additionally, the histogram in \autoref{fig:UNARTE_warmup_res5} displays a symmetric distribution of residuals centered around zero.

\begin{figure}[ht]
    \centering
    \includegraphics[width=\linewidth]{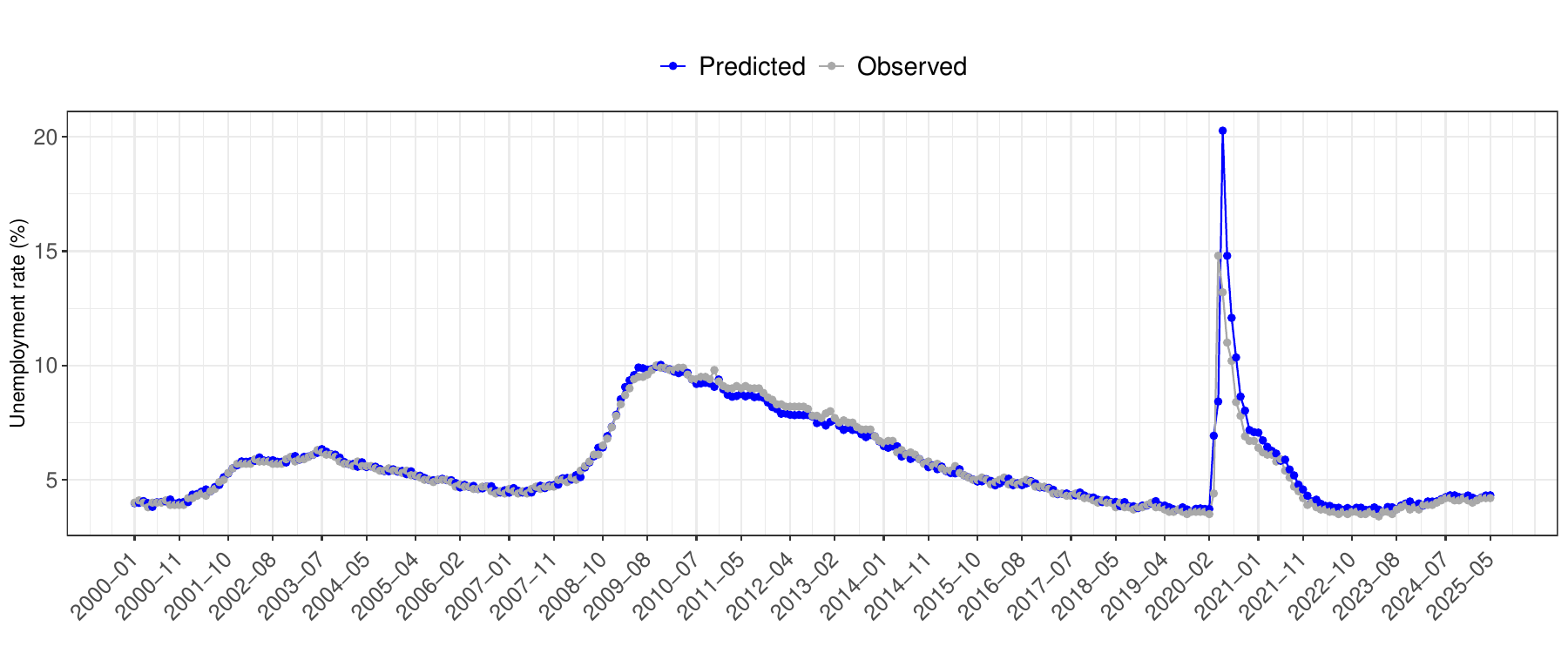}
    \caption{U.S. unemployment rate spanning from January 2000 to May 2025. Predicted values (blue) and actual values (gray)}
    \label{fig:UNRATE_realtime}
\end{figure}

\begin{figure}{
    \centering
    \raisebox{-\height}{%
        \begin{subfigure}{0.45\linewidth}
            \includegraphics[width=\linewidth]{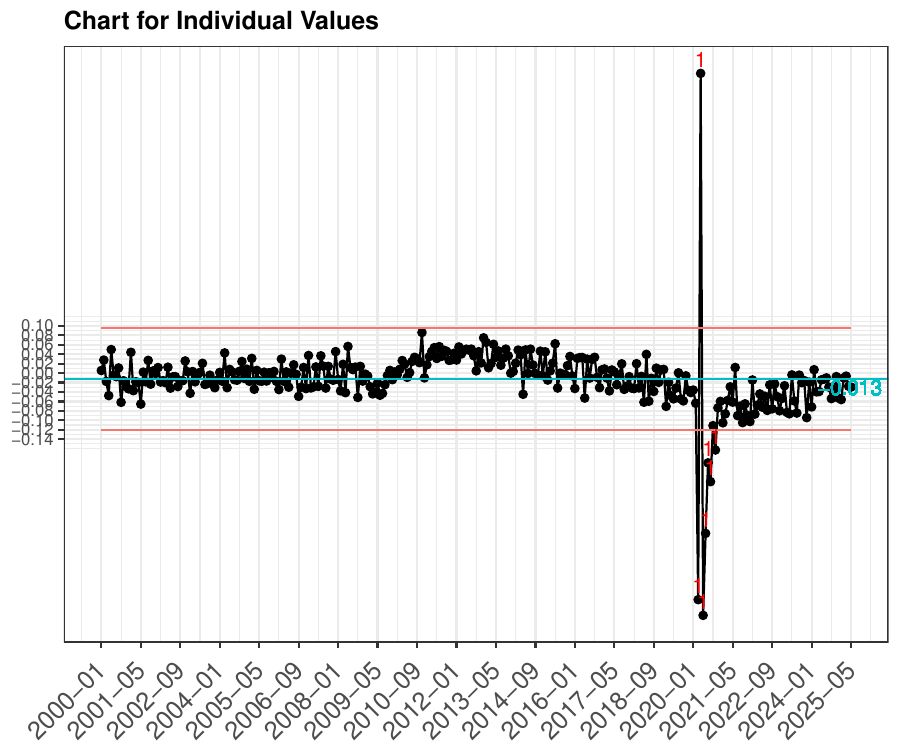}
        \end{subfigure}
        }
    \raisebox{-2.5\height}{
        \begin{subfigure}{0.45\linewidth}
            \begin{tabular}{p{4cm} c}
                \toprule
                 &  Values\\ \midrule
                Upper Limit &  0.0954 \\
                Central Line & -0.0130 \\
                Lower Limit & -0.1213 \\
                \bottomrule
            \end{tabular}
        \end{subfigure}
    }
    \subcaption{Individual residual values chart}
    \label{fig:UNARTE_realtime_res1}
    \begin{subfigure}{0.45\linewidth}
    \includegraphics[width=\linewidth]
    {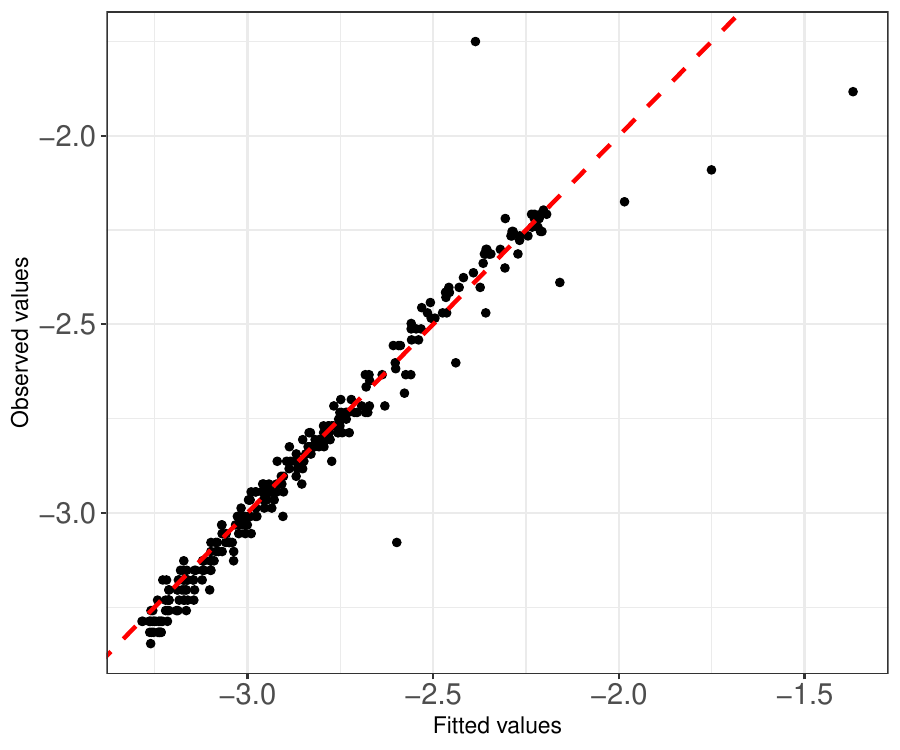}
    \caption{Observed v.s. fitted values}
    \end{subfigure}
    \begin{subfigure}{0.45\linewidth}
        \includegraphics[width=\linewidth]{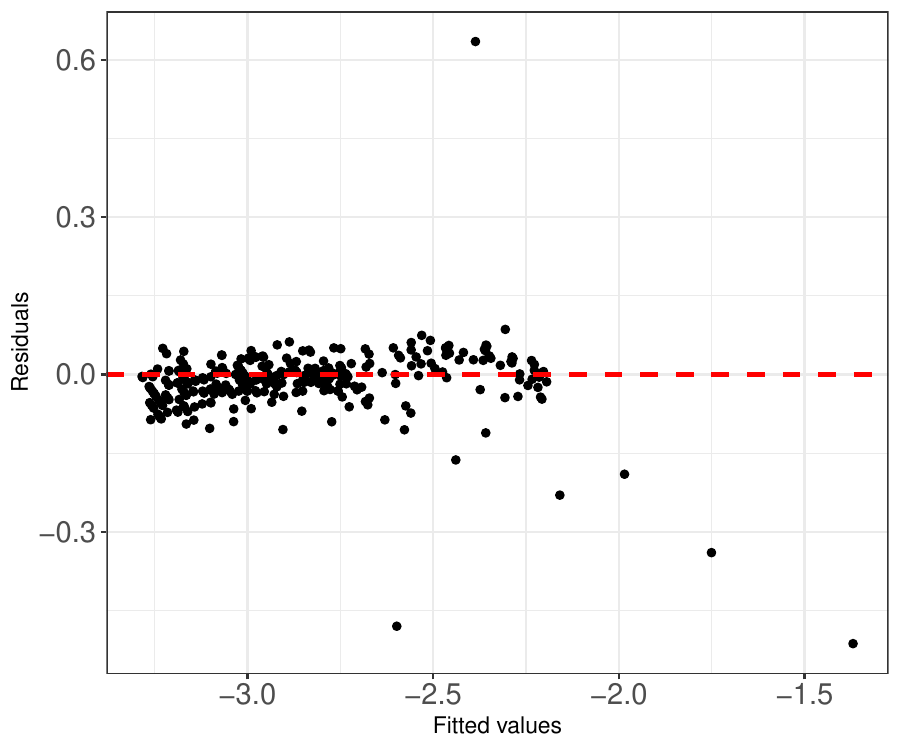}
        \caption{Residuals plot}
        \label{fig:UNARTE_realtime_res4}
    \end{subfigure}
    \raisebox{-\height}{%
        \begin{subfigure}{0.45\linewidth}
            \includegraphics[width=\linewidth]{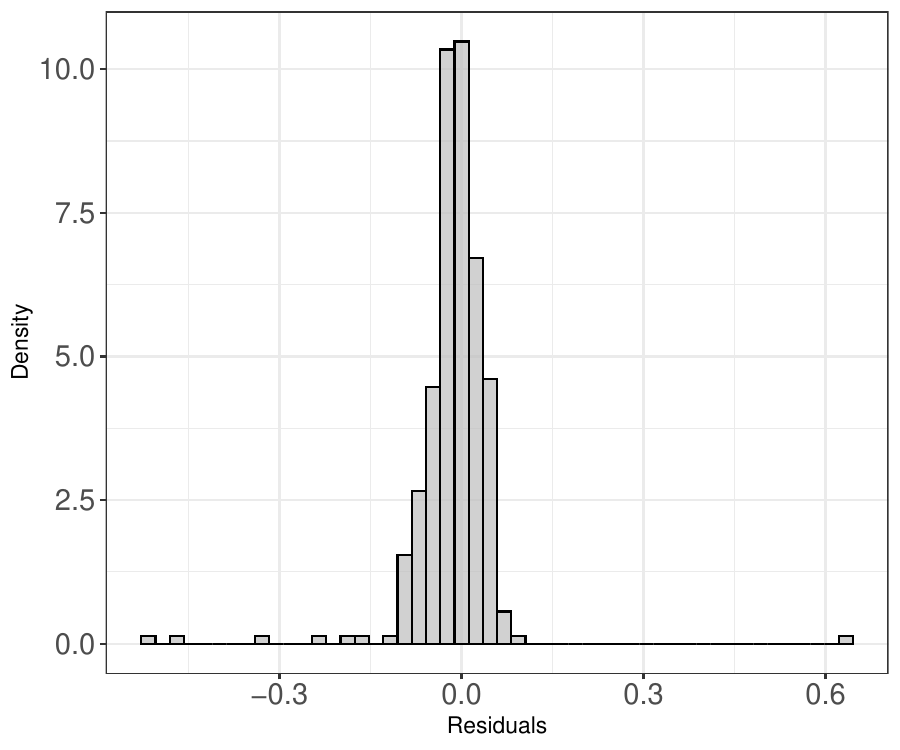}
        \end{subfigure}
    }
    \raisebox{-1.5\height}{
        \begin{subfigure}{0.45\linewidth}
            \begin{tabular}{p{4cm} c}
                \toprule
                 &  Values\\ \midrule
                Min.    & -0.5138\\
                1st Qu. & -0.0319\\
                Median  & -0.0085\\
                Mean    & -0.013\\
                3rd Qu. & 0.0159\\
                Max.    & 0.6357\\
                \bottomrule
            \end{tabular}
           \end{subfigure}
    }
    \subcaption{Exploratory of residuals}
    \label{fig:UNARTE_realtime_res5}
    }
    \caption{Residuals analysis for the real-time learning scenario}
    \label{fig:UNRATE_realtime_residuals}
\end{figure}

Utilizing the final estimated model from the warm-up period, we then proceeded to enter a real-time estimation scenario. From January 2000 to May 2025, data were processed incrementally. At every time step, each new incoming data dynamically updated both the standardization of the model matrix and the estimated parameters. \autoref{fig:UNRATE_realtime} illustrates that, in general, the forecasts are close to the true unemployment rates throughout the entire period. Two specific intervals were noted as exceptions. The first deviation began in 2009, coinciding with the Great Recession. The second deviation began in the first quarter of 2020, precisely corresponding with the onset of the COVID-19 pandemic. The results of the residual analysis are presented in \autoref{fig:UNRATE_realtime_residuals}.

Analysis of the individual values control chart for residuals (\autoref{fig:UNARTE_realtime_res1}) confirms that the observed prediction deviations during the COVID-19 pandemic are indeed outliers. A complete list of all the outliers identified (points that failed the first Nelson test) can be found in \autoref{tab:UNRATE_realtime_nelson}. We can see from the same figure that, starting from somewhere in 2008, the individual residual values present an increase trend, which indicates that the algorithm underestimated the unemployment rate. After a certain amount of time, the algorithm adapted to the underlying data drift, as evidenced by the gradual convergence of residuals to the central zero line. However, these slight deviations were not identified as outliers. The second deviation period, coinciding with the COVID-19 pandemic, caused a massive shock, with the unemployment rate spiking by over 10\% , from 4.4\% in March 2020 to 14.8\% in May 2020. Due to this high shock, the algorithm produced predictions that were slightly higher than usual. These two observations clearly demonstrate the adaptive capabilities of our algorithm to sudden or abrupt changes in the data. \autoref{tab:UNRATE_realtime_nelson_mu} shows the changes in coefficient values for these outlier data points.

\begin{table}[ht]
    \caption{Outliers detected by Nelson test 1 in individual residual values chart}
    \label{tab:UNRATE_realtime_nelson}
    \centering
    \begin{tabular}{ccccc}
      \toprule
      Date  & Forecast & True value & Lower limit & Upper limit \\
      \midrule
    2020-03 &  4.4 &  6.9266 &  0.9807 & 35.8646\\
    2020-04 & 14.8 &  8.4239 &  5.4525 & 12.7955\\
    2020-05 & 13.2 & 20.2683 & 14.4422 & 27.6841\\
    2020-06 & 11.0 & 14.7956 & 12.3572 & 17.6185\\
    2020-07 & 10.2 & 12.0806 & 10.4351 & 13.9451\\
    2020-08 &  8.4 & 10.3500 &  8.8846 & 12.0251\\
    2020-10 &  6.9 &  8.0244 &  7.3180 &  8.7925\\
    \bottomrule
    \end{tabular}
\end{table}
\begin{table}[ht]
    \caption{Changes of coefficients of outlier data point}
    \label{tab:UNRATE_realtime_nelson_mu}
    \centering
    \begin{tabular}{cccccc}
     \toprule
Date    &  intercept & UNEMP\_LAG1 & IC & CPI & IPI \\\midrule
2020-02 & -2.7531 & 0.2524 & 0.0281 & 0.0056 & -0.0140\\
2020-03 & -2.7537 & 0.2536 & 0.0173 & 0.0045 & -0.0147\\
2020-04 & -2.7519 & 0.2517 & 0.0550 & 0.0075 & -0.0136\\
2020-05 & -2.7528 & 0.2483 & 0.0471 & 0.0059 & -0.0143\\
2020-06 & -2.7535 & 0.2463 & 0.0438 & 0.0049 & -0.0148\\
2020-07 & -2.7539 & 0.2454 & 0.0421 & 0.0042 & -0.0152\\
2020-09 & -2.7545 & 0.2442 & 0.0406 & 0.0031 & -0.0160\\
\bottomrule
    \end{tabular}
\end{table}

\begin{figure}[ht]
    \centering
    \includegraphics[width=.8\linewidth]{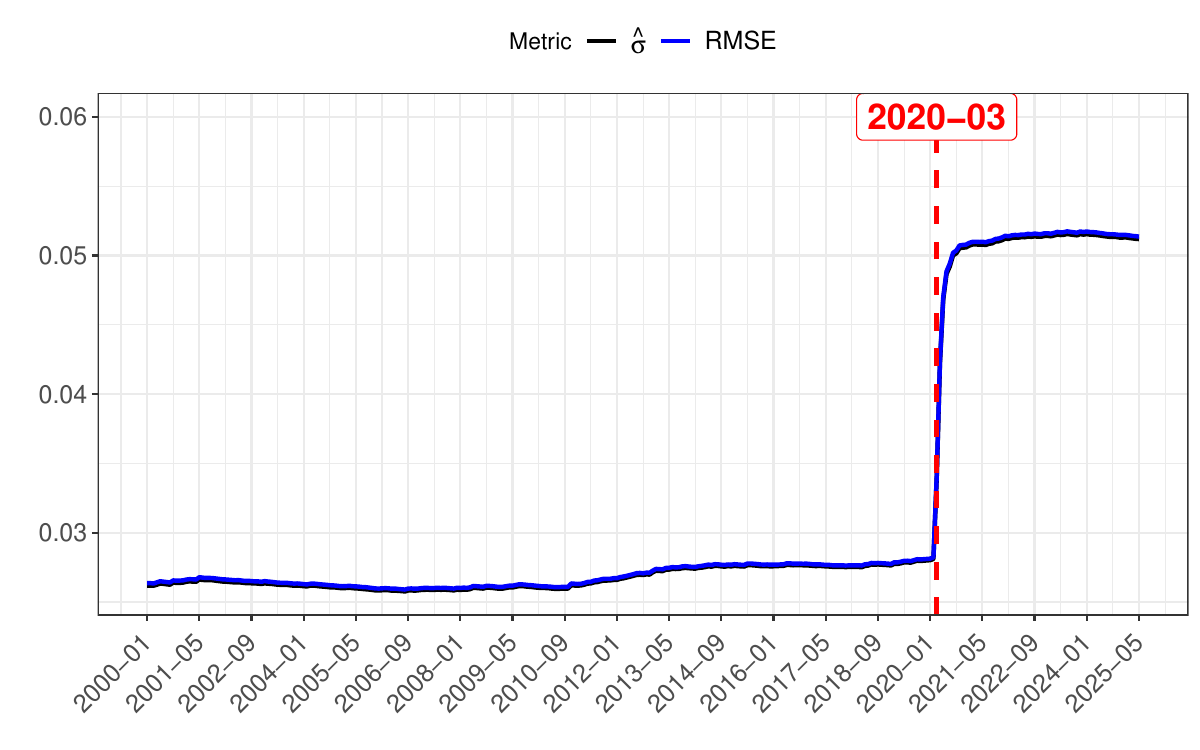}
    \caption{Metric monitoring}
    \label{fig:UNRATE_drift}
\end{figure}

The application of the first Nelson test for outlier detection suggests its efficacy in identifying early concept drift in online learning algorithms. To visualize this, \autoref{fig:UNRATE_drift} presents the evolution of $\hat{\sigma}$ and RMSE over 305 time steps. Visually, the onset of concept drift consistently aligns with the position of the first detected outlier. Detailed metrics for outlier data points are presented in \autoref{tab:UNRATE_quality2}.

\begin{table}
    \caption{Model quality measure at selected time steps in real-time learning}
    \label{tab:UNRATE_quality2}
    \centering
    \begin{tabular}{cccccccc}
    \toprule
    Date & $SST$ & $SSE$ & $n$ & $p$ & $R^2$ & $\hat{\sigma}$ & $RMSE$\\
    \midrule
2020-03 & 53.0959 & 0.7343 & 638 & 5 & 0.9862 & 0.0341 & 0.0339\\
2020-04 & 54.1529 & 1.1384 & 639 & 5 & 0.9790 & 0.0424 & 0.0422\\
2020-05 & 54.9513 & 1.4024 & 640 & 5 & 0.9745 & 0.0470 & 0.0468\\
2020-06 & 55.4205 & 1.5180 & 641 & 5 & 0.9726 & 0.0488 & 0.0487\\
2020-07 & 55.7799 & 1.5542 & 642 & 5 & 0.9721 & 0.0494 & 0.0492\\
2020-08 & 55.9279 & 1.6073 & 643 & 5 & 0.9713 & 0.0502 & 0.0500\\
2020-10 & 56.0493 & 1.6463 & 645 & 5 & 0.9706 & 0.0507 & 0.0505\\
2025-05 & 64.7679 & 1.8334 & 700 & 5 & 0.9717 & 0.0514 & 0.0512\\
    \bottomrule
    \end{tabular}
\end{table}

\autoref{tab:UNRATE_params2} shows the estimated parameter values from both the initial and final iterative steps.  The learning rate used was set to $\eta=1 \times 10^{-3}$ across all iterations. For each of the 305 time steps in the real-time forecast (from January 2000 to May 2025), we calculated a 95\% prediction level using the estimated covariance matrices of the coefficients. The empirical coverage of this prediction interval was found to be 0,9639. Out of these 305 prediction intervals, 11 failed to contain the true unemployment rates.
\begin{table}[ht]
    \centering
    \caption{Comparison of estimated parameters at the initial and final real-time learning iterative step (Feb 1967 and May 2025)}
    \label{tab:UNRATE_params2}
    \begin{tabular}{c|cccccc}
    \toprule
    Parameter & $t$ & intercept & UNEMP\_LAG1 & IC & CPI & IPI \\
    \midrule
    \multirow{2}{*}{$\hat{\mu}_t$}        & 1   & -2.7536 & 0.2461 & 0.0272 &  0.0054 & -0.0132\\
                                          & 700 & -2.7603 & 0.2495 & 0.0411 & -0.0081 & -0.0232\\
                         \midrule
    \multirow{2}{*}{$\widehat{\Sigma}_t$} &1    & 0.0064 & 0.0227 & 0.0204 & 0.0986 & 0.0968\\
                                          & 700 & 0.0051 & 0.0008 & 0.0018 & 0.0535 & 0.1299\\
    \bottomrule
    \end{tabular}
\end{table}

\subsection*{Experiment 5}

In this experiment, we applied the randomized SINDy algorithm to the Elec2 dataset, a real-world binary classification problem. The objective of the dataset focuses on predicting the direction of electricity price changes (up or down) in New South Wales (NSW), Australia. Comprising 45,312 instances, the dataset includes five features and spans the period from May 7, 1996 to December 5, 1998. \autoref{tab:Elec2_vars} provides a brief description of the variables and further details can be found in \cite{Harries}. The dataset was sourced from Kaggle (\url{https://www.kaggle.com/datasets/yashsharan/the-elec2-dataset}). All features, with the exception of ``day", were pre-normalized to the range of 0 and 1 upon download. To maintain a uniform data scale, the ``day" variable was subsequently normalized using min-max scaling. The class label distribution is shown in \autoref{tab:Elec2_class}.

\begin{table}[ht]
    \centering
    \caption{Variable descriptions for Elec2, data covering }
    \label{tab:Elec2_vars}
    \begin{tabular}{c|c}
    \toprule
    Variable&  Description\\
    \midrule
         class  &  NSW price higher (UP-1) or lower (Down-0) than 24 hour average \\
         day    &  Day of week \\
         period   &  Time (based on 30-minute periods) \\
         nswdemand & NSW electricity demand \\
         vicdemand & Victoria electricity demand\\
         transfer  & Scheduled electricity transfer between states \\
    \bottomrule
    \end{tabular}
\end{table}
\begin{table}[ht]
    \centering
    \caption{Class label distribution}
    \label{tab:Elec2_class}
    \begin{tabular}{c|cc}
        \toprule
         & Up (1) & Down (0)\\
        \midrule
        Count & 19237 & 26075 \\
        Proportion & 0.5755\% & 0.4245\% \\
        \bottomrule
    \end{tabular}
\end{table}

We utilized the initial ten thousand instances for warm-up estimation in order to determine the hyperparameter and the initial guess values for the weights in the logistic function. This procedure involved 25 train-validation set splits. The first split began with 1344 instances in the training set used to do a one-week-ahead prediction corresponding to $7 \times 48 = 336$ instances. The optimal regularization term, $\lambda = 0.8276$, was chosen based on the highest accuracy achieved across all 25 validation sets. Subsequently, a regularized logistic regression was pre-run with the chosen lambda to obtain the initial weight values for the logistic function. With these initial weight values, we then applied the proposed algorithm, updating the weights at each time step. The estimated parameters of both the initial and final iterative steps are presented in \autoref{tab:Elec2_params}.

\begin{table}[ht]
    \centering
    \caption{Estimated parameters at initial and final iterative step of the warm-up estimation period}
    \label{tab:Elec2_params}
    \begin{tabular}{c|ccccccc}
    \toprule
    Parameter & $t$ & intercept & day & period & nswdemand & vicdemand &  transfer\\
    \midrule
    \multirow{2}{*}{$\hat{\mu}_t$} & 1         & -0.0493 &  0.0912 &  0.0426 &  0.7496 &  0.1292 &  0.1267\\
                & 45312       & -0.3103 & -0.3103 & -0.3113 & -0.0876 & -0.1101 & -0.074\\
    \bottomrule
    \end{tabular}
\end{table}

\begin{figure}[ht]
     \raisebox{-\height}{%
        \begin{subfigure}{0.45\linewidth}
            \includegraphics[width=\linewidth]{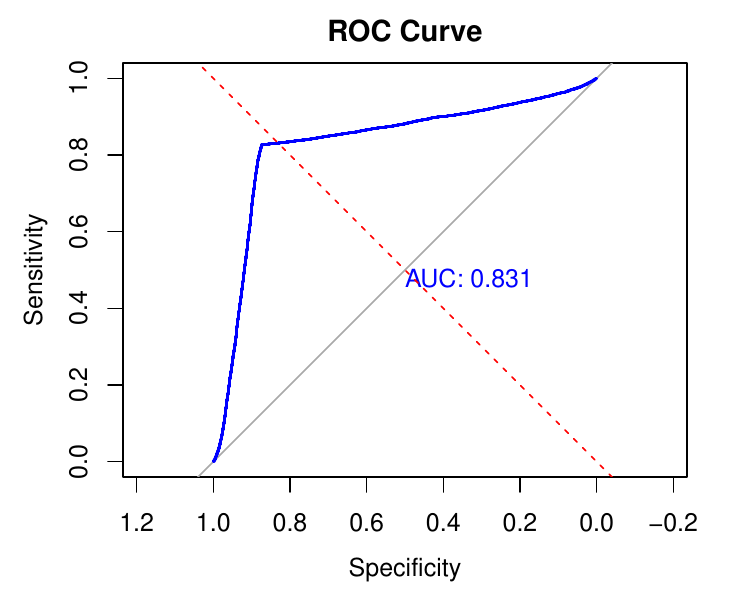}
            \caption{ROC curve}
            \label{fig:Elec2_roc}
        \end{subfigure}
     }
     \raisebox{-1.2\height}{
        \begin{subfigure}{0.45\linewidth}
            \begin{tabular}{p{4cm} c}
                \toprule
                Metric &  Values\\ \midrule
                Accuracy    & 0.8533\\       
                TPR         & 0.8272\\
                TNR         & 0.8726\\
                Precision   & 0.8273\\
                F1          & 0.8272\\
                Entropy     & 1.5377\\
                \bottomrule
            \end{tabular}
            \vspace{1cm}
            \caption{Metrics}
            \label{tab:Elec2_metric}
        \end{subfigure}
     }  
    \caption{Model performance}
    \label{fig:Elec2_performance}
\end{figure}

Finally, the classification threshold was set to 0.5654, which was derived from the ROC curve (AUC 0.8313), in \autoref{fig:Elec2_roc}. The accuracy and other performance metrics are presented in \autoref{tab:Elec2_metric}. The ROC curve shows that our model has strong predictive power with high accuracy (above 80\%).

\begin{figure}
    \centering
    \includegraphics[width=\linewidth]{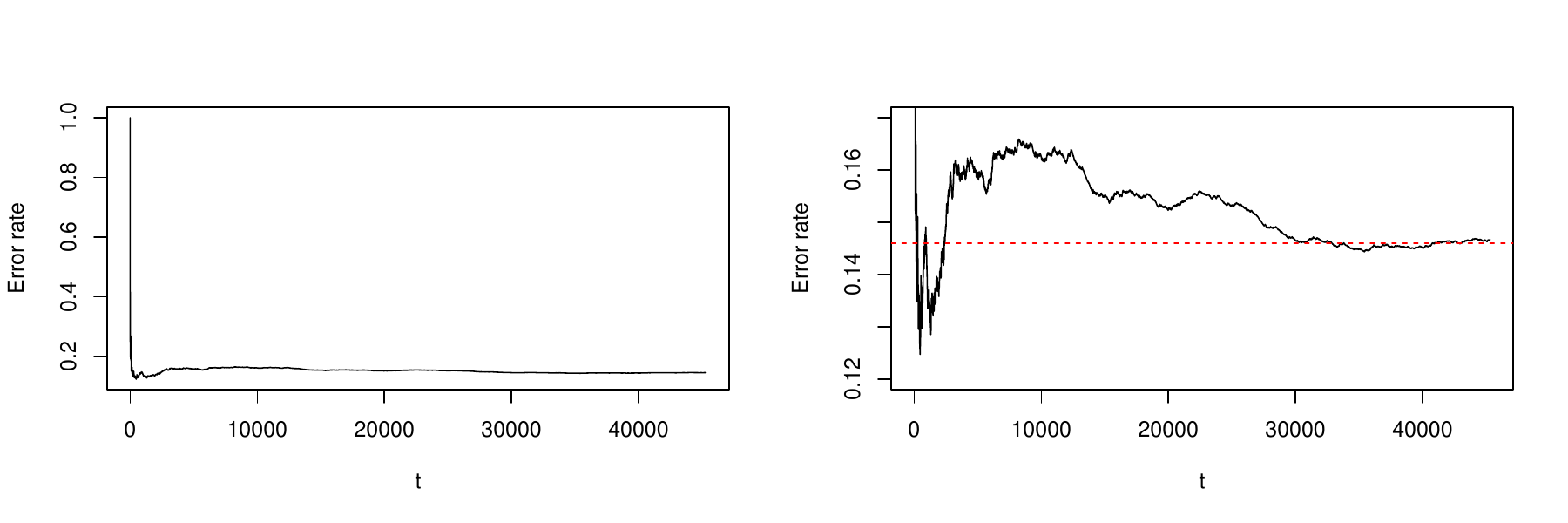}
    \caption{Error rate tracing: y-axis spanning the full range from 0 and 1 (left); y-axis zoomed into the range from 0.12 to 0.16(right)}
    \label{fig:Elec2_drift}
\end{figure}

The Elec2 dataset is widely recognized as a benchmark for concept drift detection in the online learning literature. Following the Drift Detection Method (DDM) proposed by \cite{Elec2Drift}, the trace of model's error rate, $p_i$, the probability of a false prediction up to step $i$), is presented in \autoref{fig:Elec2_drift}. The left panel of \autoref{fig:Elec2_drift} shows no apparent major changes in the overall error rate. The right panel, a zoomed-in view focusing on the range of 0.12 to 0.16, reveals a slight increase in the error rate within an early window. However, after this period, the error rate consistently decreases and then stabilize, fluctuating around a level of 0.146 indicated by the red dashed line. This observed error trace differs from those presented in the study by \cite{Elec2Drift} for the same dataset, which shows various moments of abrupt drifts. This discrepancy may be attributed to the dynamic and adaptive learning capabilities of our proposed model, highlighting its ability to adjust to changes in the data distribution even when resources for explicit major drift alerts are unavailable.

\section{Discussion and Conclusion}\label{sec:conclusion}

In this paper, we propose a randomized sequential machine learning algorithm capable of handling dynamic data structures that does not rely on the classical iid assumption. Section 2 provides an introduction to the dynamic data arrival structure from which the algorithm was developed from a probabilistic perspective. We also provide in this section an extensive proof that mathematically demonstrates the proposed algorithm satisfies the PAC learning property. Additionally, a proof-of-concept was presented in the form of a theorem and a related proof. The proposed randomized algorithm considers a class of hypotheses or predictors that fulfill a probability density function. The learner takes action dynamically at every time step according to this density. More precisely, the learner uses the probability distribution of the predictors learned from the previous time step to make a prediction for the current time step and updates the density after the incoming data for the current time step is revealed. At each time step, the algorithm  updates the weights assigned to the predictors using a gradient descent and a proximal algorithm, the latter being which is equivalent to a projection operator and ensures that the distributed weights form a valid probability density.

We were inspired by the SINDy (Sparse Identification of Nonlinear Dynamics) algorithm, a widely used data-driven modeling method for dynamics systems in recent years. We named our proposed algorithm  ``randomized SINDy" which considers the possibility of augmenting independent variables. This augmentation is sometimes referred to as dictionary or feature library, resulting in a high-dimensional model matrix that admits a sparse representation. We introduced Tikhonov regularization to the objective function to overcome the problem of ill-conditioned data. Assuming the weights follow a multivariate normal distribution, the proximal step is unnecessary. Thus, the learner's task is to estimate the parameters of the multivariate normal distribution. We explored two common applications, regression problem and binary classification with a logistic model, both of which include a regularized term in the objective function. 

Section 3 presents five experiments using simulated data and application on real-world data. The results of these experiments indicate the usefulness of our proposed algorithm. 

In conclusion, we present some observations and ideas for future studies, which are listed below.

\subsection{A smarter way to estimate an optimal learning rate and strategies for initial guess}
The learning rate was determined through empirical testing during our experimentation. It is essential to seek a more intelligent alternative to identify an optimal learning rate and hyperparameter. Additionally, the convergence rate of the algorithm depends on the initial value. How should a strategy be defined to obtain these initial guesses? How can a reasonable value be determined for the regularized term? One approach involves using a rolling or expanding window to obtain the train-validation set splits. For each set, the regression or classifier is applied to the training set and predictions are made on the corresponding validation set. Also, we vary the range of values for $\lambda$ in each model fitting. The optimal $\lambda$ is identified as the one that produces the most optimal overall performance metrics across the validation sets. However, there are challenging issues using this approach, such as the determination of the window size and the quantity of available data for warm-up estimation.

\subsection{Normalization procedures}
We applied standardization to model matrix in our experiments. A disadvantage of this normalization process is its data dependency. In real-time settings, data is dynamically fed, necessitating continual updates to the mean and standard deviation statistics. Alternative normalization procedures, such as Min-Max scaling or data-independent functions like tan and arc-tan, could be explored. 

\subsection{Seeking for a better approach to derive the update step for covariance matrix in classification}
The logistic model parameter update steps in Example 2 of Section 4 simplify the estimation procedure, as there is no exact derivative form. These simplifications for computational tractability may underestimate variation of the coefficients. 

\subsection{Computational cost}
The proposed algorithm aims to provide a scalable structure that can efficiently handle large volumes  of data. Therefore, it is important to conduct a simulation study and provide an estimation of the computation cost in terms of time consumption and memory usage.

\subsection{Performance comparisons}
A further study can be conducted by comparing the performance of the proposed algorithm on real-world data with that of some commonly applied classical methods.

\subsection{Strategies on concept drift detection}
As we saw in experiment 5, the individual residual values chart served as a useful tool for detecting potential data drift. Other Nelson tests could also be employed to identify anomalous trends in residuals, which may signal either data or concept drift. Further exploration into the capabilities of the proposed algorithm for detecting concept drift and formulating effective strategies is necessary.

\subsection{Study expansion}
To expand the study and include the case of infinite hypotheses and use related techniques to abstract Wiener space to build probability in the predictor space. 

\clearpage
\printbibliography

\end{document}